\documentclass{article}

\usepackage{PRIMEarxiv}
\usepackage[utf8]{inputenc}
\usepackage[T1]{fontenc}
\usepackage{textcomp}
\usepackage{graphicx}
\usepackage{xcolor}
\usepackage{hyperref}
\usepackage{url}
\usepackage{booktabs}
\usepackage{array}
\usepackage{amsmath}
\usepackage{amssymb}
\usepackage{nicefrac}
\usepackage{microtype}
\usepackage{placeins}

\graphicspath{{media/}}
\definecolor{linkblue}{rgb}{0,0,0.45}
\hypersetup{colorlinks=true,citecolor=linkblue,linkcolor=linkblue,urlcolor=linkblue}

\title{XRF-to-Optical Field-of-View Localization with Vision Language Models}

\author{
  Xiangyu Yin$^{1,*}$, Tatjana Paunesku$^{2}$, Letonia Copeland-Hardin$^{2,3}$,
  Martina Ralle$^{4}$, \\[0.25em]
  \textbf{Zichao Wendy Di$^{1}$, Si Chen$^{1}$, Gayle E. Woloschak$^{2}$,
  Barry Lai$^{1}$, Mathew J. Cherukara$^{1,\dagger}$, and Stefan Vogt$^{1}$} \\[0.6em]
  $^1$Advanced Photon Source, Argonne National Laboratory, Lemont, Illinois, USA \\
  $^2$Northwestern University, Chicago, Illinois, USA \\
  $^3$University of Chicago, Chicago, Illinois, USA \\
  $^4$Oregon Health and Science University, Portland, Oregon, USA \\
  $^*$\texttt{xyin@anl.gov}; $^\dagger$\texttt{mcherukara@anl.gov}
}

\begin{document}
\pagenumbering{gobble}
\thispagestyle{empty}
\textbf{GOVERNMENT LICENSE}

The submitted manuscript has been created by UChicago Argonne, LLC, Operator of Argonne National Laboratory (“Argonne”). Argonne, a U.S. Department of Energy Office of Science laboratory, is operated under Contract No. DE-AC02-06CH11357. The U.S. Government retains for itself, and others acting on its behalf, a paid-up nonexclusive, irrevocable worldwide license in said article to reproduce, prepare derivative works, distribute copies to the public, and perform publicly and display publicly, by or on behalf of the Government. The Department of Energy will provide public access to these results of federally sponsored research in accordance with the DOE Public Access Plan. \href{http://energy.gov/downloads/doe-public-access-plan}{http://energy.gov/downloads/doe-public-access-plan}
\clearpage
\pagenumbering{arabic}

\maketitle

\begin{abstract}
Registering images acquired with different microscopy modalities is essential for relating complementary measurements of the same specimen. In correlative X-ray fluorescence (XRF) and optical microscopy, the XRF map often covers only a small region of an optical image acquired from the same or an adjacent tissue section. Field-of-view (FOV) localization is necessary but can be difficult when appearance and structure differ across modalities. Here we evaluate training-free vision language model (VLM) localization on two datasets representing same-section high-correspondence and adjacent-section low-correspondence imaging. We test unconstrained and metadata-constrained search and compare VLMs with geometric controls, classical template matching, and two alternative training-free approaches (DINOv2 and multiGradICON). Direct VLM prompting produced content-dependent spatial signals but was not reliable alone. Classical matching was most accurate when cross-modal structure was preserved but failed in the low-correspondence collection. A proposal-and-verify workflow used repeated VLM predictions as candidates and image-based similarity to select the final location. This workflow recovered useful localization in the low-correspondence regime. 
\end{abstract}


\section{Introduction}
\label{sec:introduction}

Correlative microscopy can reveal information that a single modality cannot show alone~\cite{caplan2011correlative,walter2020correlative}. One informative pairing combines optical microscopy, which provides morphological context, with X-ray fluorescence (XRF) microscopy, which maps elemental distributions at resolutions from micrometers to tens of nanometers~\cite{webb2022xrf,pushie2014xrf,fahrni2007biological,dejonge2010xrf}. This combination can reveal trace-element distributions in neurons~\cite{gustavsson2021photothermal}, cellular immunofluorescence~\cite{mcrae2006microxrf,ortega2009biometals}, nanoparticle distributions under room-temperature and cryogenic conditions~\cite{paunesku2003biology,yuan2013epidermal}, and metal localization under cryogenic conditions~\cite{ortega2024cryo}. As synchrotron XRF advances in resolution and throughput, reliable correlative workflows increasingly depend on fast and robust ways to connect XRF maps back to optical images.

The prerequisite for any such analysis is XRF-to-optical localization: determining where the XRF FOV lies within the optical reference image. Optical and XRF images differ in contrast mechanism, spatial resolution, and signal-to-noise ratio. In same-section imaging, both modalities observe the same physical section, although acquisition and handling can alter its appearance. In serial-section imaging, optical and XRF measurements come from adjacent sections of the same specimen, so tissue boundaries and internal structures can differ physically. The XRF map also often covers only a small subregion of the optical image. Structures that a trained scientist can recognize across modalities may therefore share little pixel-level similarity.

Classical localization methods remain the natural first tool when the two modalities share enough structural correspondence. Fiducial markers and etched grids can provide cross-modal reference points~\cite{mohammadian2019fiducial,sheriff2021autocrim,arigundiya2026hierarchical}, but they can be costly to implement, require prospective planning, and cannot be applied retrospectively. Mutual information and gradient-based similarity measures~\cite{maes1997mutual,pluim2000gradient,pluim2003survey} can tolerate cross-modal intensity differences to some extent but can fail when the XRF target is small, contrast is sparse, or shared structure is weak. Software such as the MicroAnalysis Toolkit (SMAK) exists to support analysis of compatible multimodal images~\cite{webb2011smak}. In coordinated acquisitions, recorded stage positions can further constrain image placement~\cite{preibisch2009stitching,chalfoun2017mist}. These advantages depend on planned acquisition and do not directly extend to independently imaged adjacent sections. 

Machine learning and deep learning methods address this challenge by learning representations that can bridge appearance gaps. Examples include ROI-selection models for XRF workflows~\cite{chowdhury2022roi}, unsupervised multimodal registration~\cite{grexa2024supercut}, and contrastive multimodal representations~\cite{breznik2024crossmodality}. These approaches can work well, but they also introduce a practical barrier for scientific users: a new specimen type, beamline configuration, or modality pairing may require task-specific training data, domain adaptation, model selection, and engineering effort. A trained model can perform well within its training distribution while being difficult to repurpose for another correlative experiment or to insert into a broader automated workflow.

Recent advancements in foundation models motivate an alternative strategy. A pretrained model can perform well off-the-shelf without task-specific retraining. For example, vision language models (VLMs) accept joint image and text inputs~\cite{radford2021clip,alayrac2022flamingo}. They can use in-context visual examples and natural-language instructions, making them compatible with automated workflows in which localization is one step of a larger task. We also compare VLMs with two alternative training-free foundation model strategies: vision foundation model features from DINOv2~\cite{oquab2024dinov2} and transfer of a pretrained multimodal registration foundation model, multiGradICON~\cite{demir2024multigradicon}. We evaluate all training-free strategies alongside geometric controls and classical template matching under two settings: unconstrained search estimates the XRF FOV from the image pair alone; metadata-constrained search where preprocessing fixes orientation and the supplied box size fixes scale.

Because recent evaluations of foundation models in microscopy and materials science show that these models can remain limited in expert scientific image analysis and spatial reasoning~\cite{verma2026vlm,alampara2025probing}, we developed a proposal-and-verify workflow where VLMs are candidate generators rather than final localization oracles. The repeated VLM calls produce candidate regions, and classical image-based scores select among them.

In our computational experiments with two datasets representing same-section high-correspondence and adjacent-section low-correspondence imaging, we observe that classical matching is highly accurate when cross-modal structure is well correlated but fails on the adjacent-section collection. Direct foundation-model outputs carry useful spatial information but can be unstable and remain insufficient in both regimes. Proposal-and-verify can recover performance and convert variable VLM predictions into useful localization in the low-correspondence regime. Overall we make three specific contributions in this study: 1. We formalize XRF-to-optical field-of-view localization in unconstrained and metadata-constrained settings. 2. We compare direct VLM prompting, DINOv2 features, and multiGradICON as three training-free foundation-model strategies. 3. We introduce a proposal-and-verify workflow that treats VLM outputs as candidate regions rather than final answers.

\section{Methods}
\label{sec:methods}

\subsection{Task definition}
\label{sec:task}

We have $n$ XRF map--optical image pairs indexed by $i$. For each test case, the optical reference image $O_i$ has height $H_i$ and width $W_i$. The corresponding XRF map $X_i$ is a single-channel intensity image on its own pixel grid. It represents a FOV from the same physical section or from a serially adjacent section. The target is the XRF FOV expressed in optical-image coordinates as an axis-aligned bounding box
\[
  b_i^\ast = (x_i^\ast, y_i^\ast, w_i^\ast, h_i^\ast),
\]
where $(x_i^\ast,y_i^\ast)$ is the top-left corner and $(w_i^\ast,h_i^\ast)$ are the width and height in optical pixels. The prediction is $\hat{b}_i=(\hat{x}_i,\hat{y}_i,\hat{w}_i,\hat{h}_i)$.

\begin{figure}[!ht]
  \centering
  \includegraphics[width=0.95\linewidth]{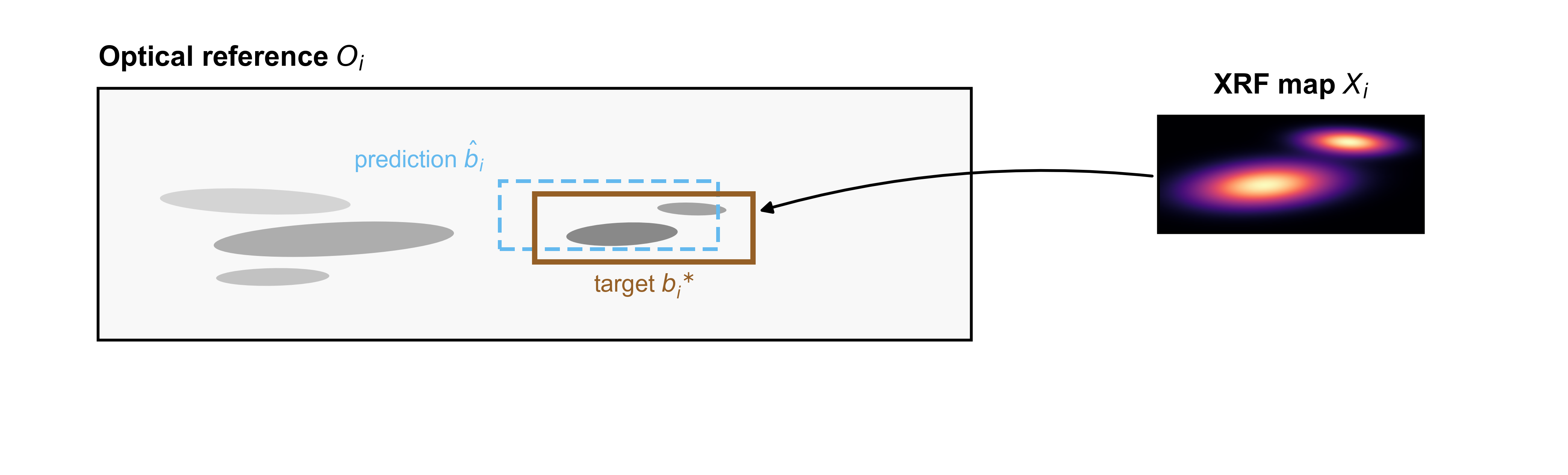}
  \caption{Task definition for XRF-to-optical FOV localization. Given an optical reference $O_i$ and an XRF map $X_i$, the goal is to estimate the XRF FOV $b_i^\ast$ in optical-image coordinates. In unconstrained search, the method estimates the full box $\hat{b}_i$. In metadata-constrained search, the supplied box size $(w_i^m,h_i^m)$ fixes scale and the method estimates only the translation $t_i$. All inputs are pre-rotated.}
  \label{fig:task_definition}
\end{figure}

We study two search settings (Fig.~\ref{fig:task_definition}). In \emph{unconstrained search}, no acquisition metadata are provided to the method, so the model must infer the full box $\hat{b}_i$ from the image pair $(O_i,X_i)$ alone. In \emph{metadata-constrained search}, the metadata supplied dimensions fix the XRF scale in optical coordinates. Translation is therefore the only estimated degree of freedom. If the constrained box size is $(w_i^m,h_i^m)$, valid top-left locations lie in
\[
  \mathcal{T}_i = [0, W_i-w_i^m] \times [0, H_i-h_i^m],
\]
and a prediction is written as
\[
  \hat{b}_i(t_i) = (t_{x,i}, t_{y,i}, w_i^m, h_i^m),
\]
where $t_i=(t_{x,i},t_{y,i}) \in \mathcal{T}_i$ is the estimated top-left location. This setting reflects experiments in which acquisition information and preprocessing constrain the pose before localization. We note that explicit rotation angles search is out of the scope of current study, thus all XRF-optical image pairs are pre-rotated to a common orientation.

\subsection{Direct vision--language localization}
\label{sec:direct_vlm}

A VLM is a multimodal foundation model that receives images and text in the same prompt and returns text. In this study, the VLM receives $O_i$, $X_i$, and a natural-language instruction asking for the XRF field-of-view location in optical-image coordinates. Few-shot prompting, also called in-context learning, places example input-output pairs in the prompt before the query case without updating the model weights~\cite{brown2020language,alayrac2022flamingo}. This makes the method lightweight to run on new specimens, because the only task-specific information supplied at inference time is the prompt, example pairs, and images.

\begin{figure}[!ht]
  \centering
  \includegraphics[width=\linewidth]{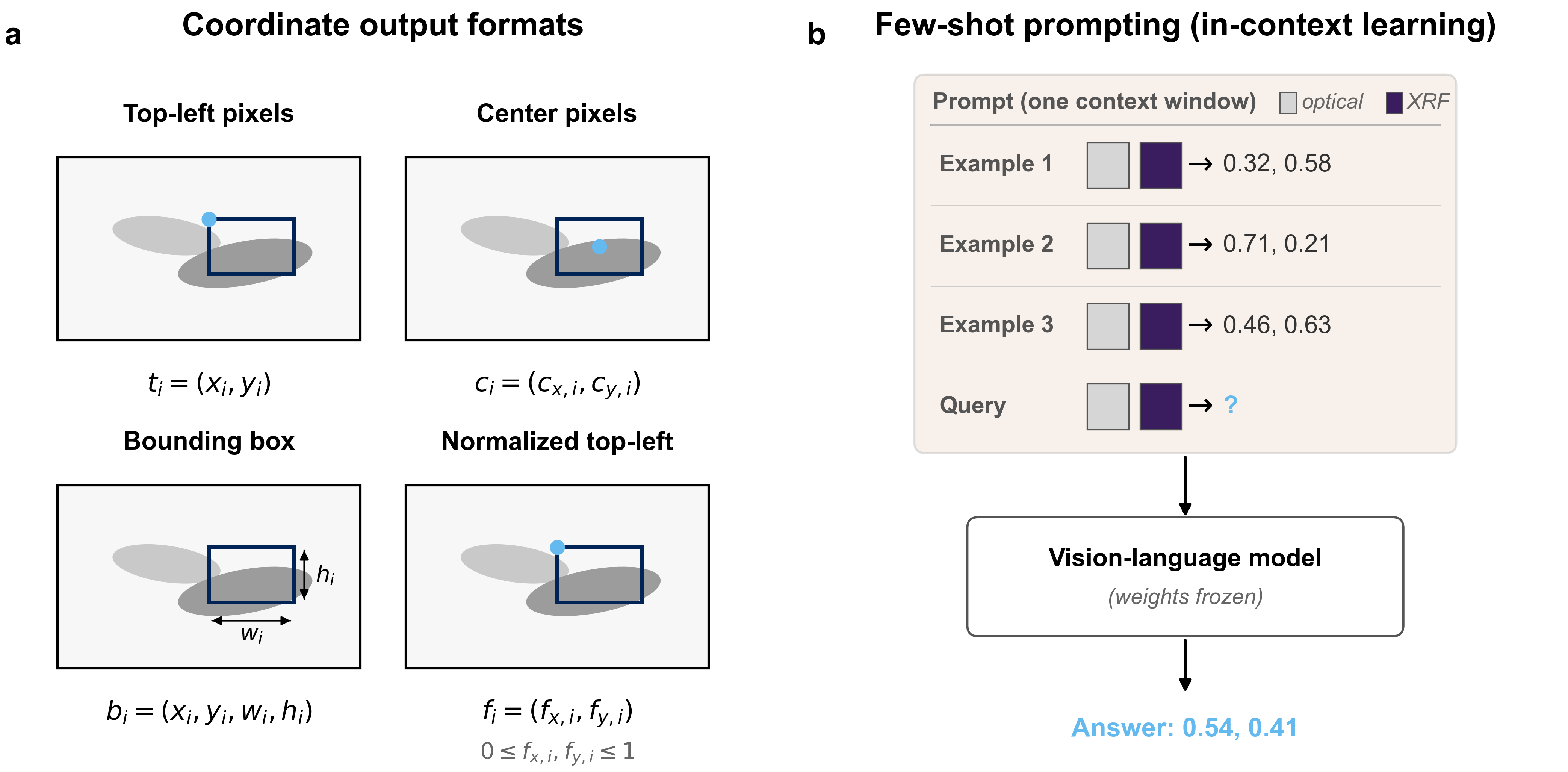}
  \caption{VLM prompting design. Four output formats were tested for direct localization: top-left pixels, center pixels, full bounding boxes, and normalized top-left fractions over the valid placement range. Few-shot prompting supplies completed examples in the prompt before the query image pair, adapting the requested response format without task-specific weight updates.}
  \label{fig:vlm_prompting}
\end{figure}

For unconstrained direct VLM, we ask for a full four-parameter box $b_i=(x_i,y_i,w_i,h_i)$. In the metadata-constrained search, instead of asking for $t_i=(t_{x,i},t_{y,i})$ directly, we ask for a normalized top-left position over the valid placement range $f_i=(f_{x,i},f_{y,i})$. The returned normalized form maps to a valid same-size box by
\[
  \hat{x}_i = f_{x,i} (W_i-w_i^m), \qquad
  \hat{y}_i = f_{y,i} (H_i-h_i^m),
\]
with $f_{x,i}$ and $f_{y,i}$ clipped to $[0,1]$.
We use this normalized top-left coordinate output format because this format gave the strongest result across three different VLMs in the ablation (Appendix~\ref{app:vlm_prompting}). Predictions outside the valid range were clipped to the image boundary. Each reported direct VLM result averages repeated three-shot calls with in-context examples drawn from other pairs. The main experiments used the GPT-5 model family through Argonne's institutional OpenAI-compatible endpoint (Argo). Appendix~\ref{app:vlm_prompting} describes the model comparison and prompting details. Figure~\ref{fig:vlm_prompting} summarizes the tested output formats and few-shot prompting design.

\subsection{Baselines and controls}
\label{sec:reference_methods}

We evaluated four classes of baselines and controls to distinguish image-based localization from overlap caused by target size or dataset-level location priors.

\textbf{Geometric controls.} Three controls quantify how much intersection-over-union (IoU) arises from task geometry without examining image content. They distinguish content-based localization from chance overlap produced by box size or dataset-level location bias. For these fixed-size controls, let $q_i=(\bar{w}_i,\bar{h}_i)$ denote the evaluated box size and let $\mathcal{T}_i(q_i)=[0,W_i-\bar{w}_i]\times[0,H_i-\bar{h}_i]$ denote the valid translation domain. In metadata-constrained search, $q_i=(w_i^m,h_i^m)$. For the unconstrained-search diagnostic controls, $q_i=(w_i^\ast,h_i^\ast)$, so that the controls isolate location priors rather than size error. The \emph{image-center prior} places the box at $\hat{t}_i=((W_i-\bar{w}_i)/2,(H_i-\bar{h}_i)/2)$. The \emph{random-box prior} samples $\hat{t}_i$ uniformly from $\mathcal{T}_i(q_i)$ and reports expected IoU. The \emph{constant prior} places boxes at the empirical mean ground-truth center across the dataset. A method that does not exceed these controls provides limited evidence of image-based localization.

\textbf{Classical template matching.} Gradient-magnitude features and normalized cross-correlation (NCC) are used to search over $\mathcal{T}_i$ at the metadata-derived box size. This is a favorable setting for classical methods when structural correspondence is strong, especially under the metadata-constrained setting because scale and orientation are already constrained and only translation is estimated.

\textbf{Self-supervised vision features.} DINOv2~\cite{oquab2024dinov2} features extracted from $O_i$ and $X_i$ localize the XRF map via feature similarity across candidate optical regions. This tests whether a general-purpose vision foundation model can bridge the optical/XRF appearance gap without additional training.

\textbf{Pretrained multimodal registration.} multiGradICON~\cite{demir2024multigradicon}, a multimodal medical-image registration network, was adapted to the optical/XRF localization setting by resizing $X_i$ to the metadata-derived size and placing it in an optical-sized canvas. This is another no-training strategy control. Its performance should be interpreted as a lower bound for a model trained or adapted specifically on optical/XRF pairs.

Baselines and controls details are summarized in Appendix~\ref{app:baselines_controls}.

\subsection{Proposal-and-verify workflow}
\label{sec:proposal_verify_workflow}

Individual VLM predictions can be too variable to serve as standalone answers, but their errors are not purely random. The proposal-and-verify workflow exploits this behavior by separating candidate generation from candidate selection. For each pair $i$, repeated VLM calls produce a candidate set
\[
  \mathcal{C}_i = \{b_{i1}, b_{i2}, \ldots, b_{iK_i}\},
\]
where each $b_{ik}$ is a parsed field-of-view proposal in optical coordinates. The final estimate is selected by an image-based score,
\[
  \hat{b}_i = \arg\max_{b \in \mathcal{C}_i} s(O_{i,b}, X_i),
\]
where $O_{i,b}$ is the optical crop defined by candidate box $b$, resized when needed to match the XRF map, and $s(\cdot)$ is a similarity measure between the optical crop and XRF map.

\subsection{Datasets and experimental settings}
\label{sec:datasets_settings}

We evaluate the methods of Sections~\ref{sec:direct_vlm}--\ref{sec:proposal_verify_workflow} on two independent pair collections. The \emph{controlled} collection contains $n{=}5$ same-section pairs with high cross-modal correspondence. The \emph{challenging} collection contains $n{=}9$ adjacent-section pairs with low correspondence. The controlled collection comprises one optical reference and five XRF regions of interest from the same physical specimen section. All were acquired at beamline 2-ID-D of the Advanced Photon Source (APS), using 81$\times$81 or 91$\times$91 scan grids with a 10\,\textmu{}m step size. The challenging collection comprises four specimen groups. Each group pairs an H\&E optical image with XRF maps of a serially adjacent section from the same tissue block. Sample preparation followed the procedure described by Copeland-Hardin et al.~\cite{copeland2023proof}. Briefly, archival formalin-fixed, paraffin-embedded lymph node and lung tissues from beagle dogs were obtained from the Northwestern University Radiobiology Archive (NURA\footnote{https://sites.northwestern.edu/nura/data/inhalation-toxicology-research-institute-data/}). Serial sections 5--10\,\textmu{}m thick were cut with a microtome. One section was mounted on glass, stained with hematoxylin and eosin (H\&E), and imaged with a Hamamatsu NanoZoomer slide scanner. The adjacent section was mounted on an Ultralene membrane and mapped by XRF microscopy at APS beamline 2-ID-E. Experts established ground-truth locations from tissue and void patterns together with sample-trimming information. No fiducial markers were used.

For the XRF input we use the phosphorus (P) channel as the single-channel intensity image throughout. Phosphorus is present in essentially all biological specimens and produces a dense, reliable signal across the specimen types studied here. Channel-selection sensitivity is not the focus of this study and is left to future work.

\subsection{Evaluation and statistics}
\label{sec:statistics}

For a single prediction $\hat{b}_i=(\hat{x}_i,\hat{y}_i,\hat{w}_i,\hat{h}_i)$, the localization accuracy is measured by intersection-over-union (IoU),
\[
  \mathrm{IoU}(\hat{b}_i, b_i^\ast)
  =
  \frac{\mathrm{area}(\hat{b}_i \cap b_i^\ast)}
       {\mathrm{area}(\hat{b}_i \cup b_i^\ast)}.
\]

For each setting and method, we report mean IoU,
\[
  \overline{I} = \frac{1}{n}\sum_{i=1}^n \mathrm{IoU}(\hat{b}_i,b_i^\ast),
\]
supplemented by median IoU, bootstrap 95\% confidence intervals, and pair-level values. Confidence intervals are computed by resampling pairs with replacement. 

\section{Results}
\label{sec:results}

\subsection{Localization performance across regimes and search settings}
\label{sec:reference_results}

Table~\ref{tab:search_results} reports every method on both datasets and in both settings. Under metadata-constrained search, classical template matching achieved a mean IoU of 0.92 [0.89, 0.95] on the controlled collection but only 0.01 [0.00, 0.02] on the challenging collection. Proposal-and-verify achieved 0.20 [0.05, 0.35] on the controlled collection and 0.35 [0.14, 0.58] on the challenging collection. Thus, the leading method depends on the physical correspondence between the image pair rather than on one method being universally superior.

Without the supplied box size, the same gradient+NCC pipeline failed on both collections (Table~\ref{tab:search_results}, unconstrained columns). Constraining the box size also increased direct VLM mean IoU from 0.05 to 0.10 on controlled and from 0.11 to 0.19 on challenging. For classical matching on controlled, it increased mean IoU from 0.12 to 0.92. These comparisons show that pre-rotation and scale information materially simplify localization.

Proposal-and-verify shows the same dependence on the search constraint. Under unconstrained search, it reached 0.09 [0.00, 0.19] on controlled and 0.19 [0.05, 0.39] on challenging. Under metadata-constrained search, it verified translation-only proposals at the supplied box size and reached 0.20 [0.05, 0.35] and 0.35 [0.14, 0.58], respectively. We note that the mean was higher than direct prompting in every cell.


Pretrained foundation-model controls (DINOv2 0.20, multiGradICON 0.19) and direct VLM prompting (0.19) reach similar mean IoU on the challenging dataset under metadata-constrained search, all near the strongest geometric controls. This does not imply that these methods could not perform better after task-specific adaptation. This shows that off-the-shelf pretrained representations do not automatically solve this localization problem without further training or a workflow that constrains and verifies their outputs.

\begin{figure}[!ht]
  \centering
  \includegraphics[width=0.95\linewidth]{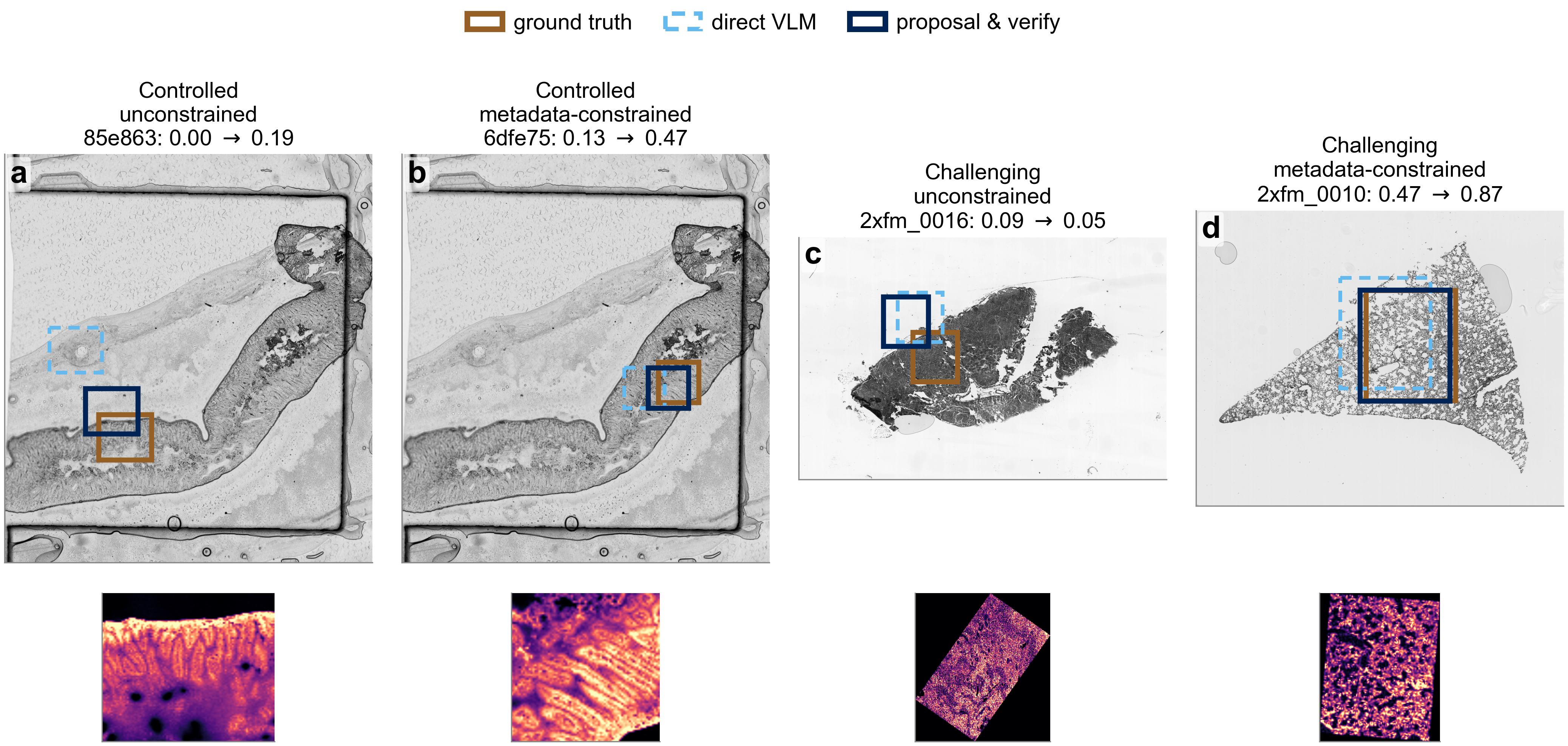}
  \caption{Representative outcomes from the four dataset--setting cells, including improvements and one small regression under challenging unconstrained search. Top: optical reference with ground truth (ochre), direct VLM prediction (dashed blue), and proposal-and-verify prediction (navy). Bottom: XRF template in false color. IoU values list direct prompting followed by proposal-and-verify. The challenging collection contains adjacent-section pairs with ground-truth FOV area below 20\% of the optical reference. All inputs are pre-rotated, so the metadata-constrained task estimates translation rather than rotation. Residual appearance differences can arise from the adjacent physical sections.}
  \label{fig:qualitative_examples}
\end{figure}

\begin{table}[!ht]
  \centering
  \scriptsize
  \caption{Localization performance (mean IoU with bootstrap 95\% confidence intervals over image pairs). Each method is reported on both datasets and in both settings. Best mean per column is bold. Geometric controls and the two pretrained baselines use the same box size in both settings because the constraint does not change their computation, so their values are repeated across the two setting columns within each dataset. Setting and policy details by cell are in Appendix~\ref{app:baselines_controls} and Appendix~\ref{app:proposal_verify_details}.}
  \label{tab:search_results}
  \begin{tabular*}{\linewidth}{@{\extracolsep{\fill}}lcccc@{}}
    \toprule
     & \multicolumn{2}{c}{Controlled ($n{=}5$)} & \multicolumn{2}{c}{Challenging ($n{=}9$)} \\
    \cmidrule(lr){2-3} \cmidrule(lr){4-5}
    Method & Unconstr. & Meta-constr. & Unconstr. & Meta-constr. \\
    \midrule
    \multicolumn{5}{@{}l}{\textbf{Geometric controls}} \\
    Image-center prior         & 0.00 [0.00, 0.00] & 0.00 [0.00, 0.00] & \textbf{0.25} [0.06, 0.48] & 0.25 [0.06, 0.48] \\
    Random-box prior           & 0.01 [0.01, 0.02] & 0.01 [0.01, 0.02] & 0.06 [0.01, 0.13] & 0.06 [0.01, 0.13] \\
    Constant prior             & 0.11 [0.00, 0.33] & 0.11 [0.00, 0.33] & 0.20 [0.03, 0.42] & 0.20 [0.03, 0.42] \\
    \addlinespace[0.25em]
    \multicolumn{5}{@{}l}{\textbf{Baselines and pretrained controls}} \\
    DINOv2 dense features      & 0.06 [0.00, 0.14] & 0.06 [0.00, 0.14] & 0.20 [0.00, 0.40] & 0.20 [0.00, 0.40] \\
    multiGradICON pretrained   & 0.00 [0.00, 0.00] & 0.00 [0.00, 0.00] & 0.19 [0.02, 0.37] & 0.19 [0.02, 0.37] \\
    Classical template matching & \textbf{0.12} [0.00, 0.36] & \textbf{0.92} [0.89, 0.95] & 0.00 [0.00, 0.00] & 0.01 [0.00, 0.02] \\
    \addlinespace[0.25em]
    \multicolumn{5}{@{}l}{\textbf{Vision--language methods}} \\
    Direct VLM, 3 examples     & 0.05 [0.00, 0.09] & 0.10 [0.04, 0.16] & 0.11 [0.04, 0.18] & 0.19 [0.06, 0.32] \\
    Proposal-and-verify        & 0.09 [0.00, 0.19] & 0.20 [0.05, 0.35] & 0.19 [0.05, 0.39] & \textbf{0.35} [0.14, 0.58] \\
    \bottomrule
  \end{tabular*}
\end{table}

\subsection{Direct prompting and the proposal-and-verify workflow}
\label{sec:direct_results}

Direct VLM prompting provides useful signal, but not enough for reliable single-pass localization. Across GPT-5, Claude Opus 4.8, and Gemma 4 31B, normalized top-left coordinates were the strongest three-example format. Direct mean IoU nevertheless remained 0.15--0.19 on the challenging metadata-constrained collection (Appendix~\ref{app:vlm_prompting}). For the selected GPT-5 configuration, mean IoU was 0.19. This exceeded the random-box prior (0.06) but was still below the image-center prior (0.25) and similar to the constant prior (0.20). Direct prompting therefore remains a candidate generator rather than a final localization method.

We then tested proposal-and-verify with the selected GPT-5 prompting configuration. On the challenging dataset under metadata-constrained search, it increased mean IoU from 0.19 to 0.35. The resulting mean exceeded the image-center prior (0.25) and classical template matching (0.01). The gain over direct prompting was nominally significant ($p{=}0.031$, exact one-sided paired sign-flip test). Full results are in Table~\ref{tab:appendix_statistical_tests}. Repeated VLM calls provide candidate fields of view, and verification selects the optical crop that best matches the XRF map. 

On the controlled dataset the workflow increases the direct VLM mean from 0.05 to 0.09 under matched unconstrained full-box prompting and from 0.10 to 0.20 under metadata-constrained search, but remains far below classical template matching with the metadata-derived box size (0.92). The two-stage procedure cannot recover when all initial proposals miss the correct region. This reinforces the two operating-regime design: proposal-and-verify is most useful where classical similarity-based localization fails, not where structural correspondence already makes classical registration decisive.

\subsection{Pair-level behavior and the proposal ceiling}
\label{sec:pair_level_results}

\begin{figure}[!ht]
  \centering
  \includegraphics[width=0.95\linewidth]{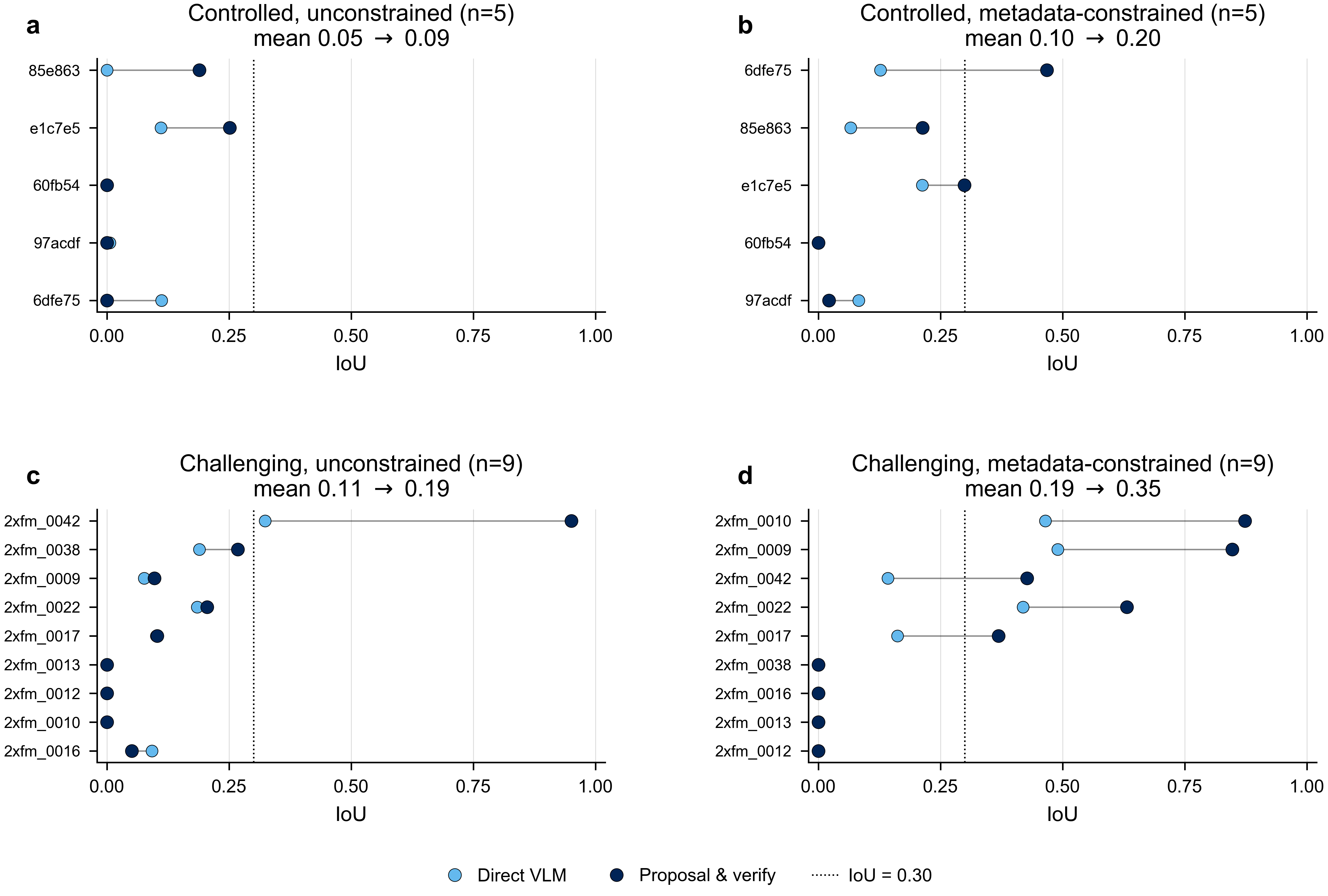}
  \caption{Pair-level comparison between direct VLM prompting and proposal-and-verify for all four dataset--setting cells in Table~\ref{tab:search_results}. Each row links the direct-prompting IoU to the proposal-and-verify IoU for one pair. The dotted line at IoU\,=\,0.30 marks the threshold above which predictions count as approximate localizations. Rows are sorted by proposal-and-verify improvement within each panel.}
  \label{fig:pair_level_behavior}
\end{figure}

We classify each prediction using two IoU thresholds. IoU below 0.10 indicates the wrong neighborhood. Values from 0.10 to below 0.30 are marginal overlaps, values from 0.30 to below 0.60 are approximate localizations, and values of at least 0.60 are good localizations. The 0.10 and 0.30 cutoffs match the wrong-neighborhood and rescue thresholds in the qualitative atlas (Fig.~\ref{fig:appendix_qualitative_atlas}). Figure~\ref{fig:pair_level_behavior} marks the 0.30 cutoff.

The pair-level view across all settings shows that the aggregate gains are not driven by a single outlier (Fig.~\ref{fig:pair_level_behavior}; Table~\ref{tab:appendix_pair_results}). Proposal-and-verify improves the mean IoU over direct prompting for most cases across all settings. We also observe that proposal-and-verify cannot rescue pairs where the candidate pool never reaches the correct neighborhood, and several small challenging target cases remain failures or even slightly worse. This reinforces our argument that VLMs should be viewed as a candidate generator rather than a trusted localization oracle. Figure~\ref{fig:qualitative_examples} includes both improvements and a small challenging-unconstrained regression case, and the full qualitative atlas across all evaluated pairs and settings is in Fig.~\ref{fig:appendix_qualitative_atlas}.

Without metadata, the system must infer both size and location before verification. Image-based scores cannot recover the correct answer if the candidate set omits it or the verifier ranks a nearby incorrect candidate higher. The metadata-constrained setting is more stable because pre-rotation and fixed scale reduce proposal variability. Leave-one-pair-out cross-validation on the challenging dataset reproduced the full-data mean IoU of 0.35, with the same scoring rule selected across folds. This indicates no detected sensitivity to one held-out pair.

\FloatBarrier

\section{Discussion}
\label{sec:discussion}

VLMs can contribute to XRF-to-optical localization as training-free candidate generators when classical similarity measures and off-the-shelf pretrained representations are weak. Classical template matching and proposal-and-verify serve complementary physical regimes. Classical matching is the appropriate first method for same-section pairs with preserved structure. VLM-guided proposal-and-verify becomes useful for adjacent-section pairs when classical scores fail, fiducials are absent, or an operator needs several candidate regions.

The comparison among DINOv2, multiGradICON, and VLM prompting exposes different transfer assumptions. DINOv2 tests whether general visual features align across optical and XRF contrast. multiGradICON tests whether a pretrained multimodal registration model transfers across domains. VLM prompting tests whether a multimodal language interface can use visual context and in-context examples to generate spatial proposals. None provided reliable direct localization without task-specific training. The three-model VLM ablation also showed sensitivity to model selection, output format and example count. The proposal-and-verify result instead supports integrating the selected VLM recipe into a search-and-selection system.

These conclusions align with the broader shift from purpose-built models toward general-purpose foundation models in scientific machine learning. For microscopy practice, the advantage of a VLM is not only that it avoids task-specific training data. The same interface can be reprompted for different coordinate systems or augmented with metadata through in-context examples~\cite{brown2020language,alayrac2022flamingo}, queried for self-reported uncertainty and failure modes~\cite{kadavath2022know}, and wired into automated or agentic scientific workflows where localization is one step in a longer experimental procedure~\cite{boiko2023autonomous,bran2024chemcrow}. These properties are difficult to capture with IoU alone, but they matter for deployment in beamline and laboratory settings.

Our findings complement XRF-to-RGB localization methods that are effective when images share structural features~\cite{amiri2025xrf,bock2023registration}. The present study addresses small-target adjacent-section pairs, where structural correspondence is weaker and training data for a dedicated model may be unavailable. In this regime, VLM proposals provide additional candidate regions when paired with independent image-based verification and operator oversight. We do not interpret this gain as evidence of intrinsic rotation or scale robustness. All primary inputs were pre-rotated, and scale was fixed in the metadata-constrained task. Current VLMs are also weak at fine-grained orientation reasoning~\cite{zhang2025spinbench}. The observed advantage is limited to generating plausible translations when low-level correspondence is insufficient.

\section{Conclusion}
\label{sec:conclusions}

XRF-to-optical field-of-view localization is difficult because the two modalities share specimen structure without reliable appearance correspondence. Direct VLM prompting is too variable for standalone localization, but its predictions contain enough spatial signal to serve as proposals. Proposal-and-verify improves the mean over direct prompting in both search settings on both datasets and reaches its highest accuracy in the low-correspondence regime. Classical template matching remains the preferred method when structural correspondence is strong. When it is weak, VLM-guided proposal-and-verify provides a training-free source of candidates for image-based selection.

\clearpage
\appendix
\counterwithin{figure}{section}
\counterwithin{table}{section}
\renewcommand{\thefigure}{\thesection\arabic{figure}}
\renewcommand{\thetable}{\thesection\arabic{table}}
\renewcommand{\theHfigure}{appendix.\Alph{section}.\arabic{figure}}
\renewcommand{\theHtable}{appendix.\Alph{section}.\arabic{table}}

\section{Dataset}
\label{app:dataset_search_settings}

Table~\ref{tab:appendix_dataset_geometry} lists the evaluated image pairs. Pair selection was fixed before evaluating any VLM-based method. The controlled dataset contains five same-section high-correspondence pairs. The challenging dataset contains nine adjacent-section low-correspondence pairs selected using a ground-truth FOV-area threshold below 20\%.

\begin{table}[!ht]
  \centering
  \footnotesize
  \caption{Geometry of evaluated image pairs. FOV area is the ground-truth XRF field-of-view area as a percentage of the optical reference image area. For adjacent-section pairs, the dimensions describe the evaluation coordinate frames and do not imply identical physical tissue outlines.}
  \label{tab:appendix_dataset_geometry}
  \resizebox{\linewidth}{!}{%
  \begin{tabular}{@{}llccc@{}}
    \toprule
    Dataset & Pair & Optical size (px) & XRF FOV size in optical pixels & FOV area (\%) \\
    \midrule
    Challenging & \texttt{2xfm\_0009} & $5120 \times 4112$ & $2013 \times 2019$ & 19.3 \\
    Challenging & \texttt{2xfm\_0010} & $5120 \times 4112$ & $1249 \times 1538$ & 9.1 \\
    Challenging & \texttt{2xfm\_0012} & $5120 \times 4112$ & $335 \times 324$ & 0.5 \\
    Challenging & \texttt{2xfm\_0013} & $5120 \times 4112$ & $261 \times 232$ & 0.3 \\
    Challenging & \texttt{2xfm\_0016} & $3840 \times 2528$ & $467 \times 515$ & 2.5 \\
    Challenging & \texttt{2xfm\_0017} & $3840 \times 2528$ & $423 \times 500$ & 2.2 \\
    Challenging & \texttt{2xfm\_0022} & $4352 \times 2400$ & $706 \times 706$ & 4.8 \\
    Challenging & \texttt{2xfm\_0038} & $4352 \times 3008$ & $87 \times 228$ & 0.2 \\
    Challenging & \texttt{2xfm\_0042} & $4352 \times 3008$ & $138 \times 237$ & 0.2 \\
    \addlinespace[0.25em]
    Controlled & \texttt{60fb54} & $5456 \times 6058$ & $732 \times 792$ & 1.8 \\
    Controlled & \texttt{6dfe75} & $5456 \times 6058$ & $600 \times 600$ & 1.1 \\
    Controlled & \texttt{85e863} & $5456 \times 6058$ & $781 \times 673$ & 1.6 \\
    Controlled & \texttt{97acdf} & $5456 \times 6058$ & $591 \times 663$ & 1.2 \\
    Controlled & \texttt{e1c7e5} & $5456 \times 6058$ & $699 \times 699$ & 1.5 \\
    \bottomrule
  \end{tabular}%
  }
\end{table}

\FloatBarrier

\section{VLM Prompting and Ablations}
\label{app:vlm_prompting}

The VLM prompting used a fixed system instruction that identifies the first image as the optical reference and the second as the XRF template. It then requests numeric output only. User prompts requested one coordinate format from Section~\ref{sec:direct_vlm} at a time. Parsed normalized outputs were clipped to $[0,1]$ before conversion to optical pixels, and pixel-coordinate outputs were clipped to the image boundary. Few-shot examples were inserted as completed image-pair and coordinate-response demonstrations before the query pair. This is in-context learning rather than fine-tuning because examples change the prompt but not the model weights. GPT-5 and Claude Opus 4.8 were accessed through Argonne's internal Argo endpoint using the model identifiers \texttt{gpt5} and \texttt{claudeopus48}, respectively. Gemma was served locally using \texttt{google/gemma-4-31b} with Q8\_0 GGUF quantization. Request payloads did not set decoding parameters, so each service used its serving defaults.

\begin{table}[!ht]
  \centering
  \scriptsize
  \caption{Direct VLM ablations on the challenging metadata-constrained collection ($n{=}9$). Each entry is the mean IoU over nine pair means, with three calls per pair (27 calls per entry).}
  \label{tab:appendix_vlm_ablations}
  \begin{tabular}{@{}p{0.15\linewidth}p{0.25\linewidth}crrr@{}}
    \toprule
    Ablation & Condition & Examples & GPT-5 & \shortstack{Claude\\Opus 4.8} & \shortstack{Gemma 4 31B\\Q8\_0} \\
    \midrule
    Output format & Pixel top-left & 3 & 0.07 & 0.06 & 0.13 \\
    Output format & Pixel center & 3 & 0.06 & 0.08 & 0.08 \\
    Output format & Normalized top-left & 3 & 0.19 & 0.17 & 0.15 \\
    \addlinespace[0.25em]
    Few-shot count & Normalized top-left & 0 & 0.18 & 0.19 & 0.12 \\
    Few-shot count & Normalized top-left & 1 & 0.12 & 0.12 & 0.10 \\
    Few-shot count & Normalized top-left & 2 & 0.17 & 0.15 & 0.11 \\
    Few-shot count & Normalized top-left & 3 & 0.19 & 0.17 & 0.15 \\
    \bottomrule
  \end{tabular}
\end{table}

\FloatBarrier

\section{Baselines and Controls}
\label{app:baselines_controls}

Table~\ref{tab:appendix_baseline_details} gives implementation details for the non-VLM methods. The two trained or pretrained baselines were not trained specifically on these optical/XRF datasets. Their limited transfer performance is therefore consistent with the manuscript's motivation: purpose-trained ML/DL methods can be powerful, but off-the-shelf transfer to a new correlative microscopy distribution is not guaranteed.

\paragraph{Classical template matching without the metadata-derived box size.} The unconstrained-setting classical entries in Table~\ref{tab:search_results} (0.12 controlled, 0.00 challenging) were obtained by replacing the fixed metadata-derived template size with a multi-scale search over short-side sizes $\{64, 96, 128, 192, 256, 384, 512, 768, 1024, 1536, 2048\}$~px. Both runs collapse because multi-scale NCC consistently selects the smallest scale in the grid, a known pathology of normalized cross-correlation under weak local intensity correspondence. This pathology is independent of the regime: it appears on the high-correspondence dataset as well, where the same pipeline reaches 0.92 once the template size is fixed.

\begin{table}[!ht]
  \centering
  \footnotesize
  \caption{Implementation details for baselines and controls.}
  \label{tab:appendix_baseline_details}
  \begin{tabular}{@{}p{0.20\linewidth}p{0.43\linewidth}p{0.28\linewidth}@{}}
    \toprule
    Method & Implementation & Role in interpretation \\
    \midrule
    Image-center prior & Places the evaluated box at the optical-image center. The box size is metadata-derived on the challenging dataset, and uses the ground-truth size on the controlled dataset where no probe metadata are recorded. The same box-size convention applies to the random-box and constant priors. & Measures how much IoU can arise from center bias alone. \\
    Random-box prior & Samples valid top-left locations uniformly and reports expected IoU from 5000 samples per pair. & Measures chance overlap under the same box-size convention. \\
    Constant prior & Places each box at the empirical mean ground-truth center across the evaluated setting. & Measures dataset-level location bias. \\
    Classical template matching & Uses gradient-magnitude features and OpenCV normalized cross-correlation over valid translations. Additional development variants included raw NCC, percentile-normalized NCC, and gradient mutual information. & Strong training-free reference when cross-modal structure is shared. \\
    DINOv2 dense features & Uses \texttt{facebook/dinov2-base}. Grayscale inputs are replicated to RGB, dense patch features are normalized, and candidate optical windows are scored by cosine similarity to the template feature. & Tests whether generic self-supervised vision features align optical and XRF appearance without task-specific training. \\
    multiGradICON pretrained & Places the resized XRF map into an optical-sized canvas and runs \texttt{unigradicon-register} with the multiGradICON model for 30 iterations. Both inputs are passed through the same medical-image registration interface. & Tests no-adaptation transfer of a pretrained multimodal registration model. \\
    \bottomrule
  \end{tabular}
\end{table}

\FloatBarrier

\section{Proposal-and-Verify Details}
\label{app:proposal_verify_details}

\paragraph{Scoring functions.} The proposal-and-verify workflow first builds a finite candidate set $\mathcal{C}_i$ from repeated VLM calls and then selects one candidate by an image-based score. For a candidate box $b$, the optical crop $O_{i,b}$ is resized to the XRF map size before scoring. The verifier records four complementary scores. Normalized mutual information (NMI) measures statistical dependence between the optical crop and XRF map,
\[
  \mathrm{NMI}(O_{i,b}, X_i)
  =
  \frac{I(O_{i,b}; X_i)}{\sqrt{H(O_{i,b})H(X_i)}},
\]
where $I(\cdot;\cdot)$ denotes mutual information and $H(\cdot)$ denotes image-intensity entropy. Percentile-normalized correlation measures linear correspondence after robust intensity normalization. Gradient correlation compares edge-like structure. Edge-chamfer similarity measures the distance between detected edge maps.

\paragraph{Policies by cell.} Different policies were selected for different combinations of dataset and search setting. On the challenging dataset under metadata-constrained search, repeated few-shot predictions form the primary proposal pool, scored with edge-chamfer similarity for sufficiently large templates and with NMI for very small templates. A supplemental zero-shot pool can override only when its top candidate both improves the image-based score and overlaps the primary candidate by at least 0.5 IoU, preventing spatially unrelated overrides. On both datasets under unconstrained search, a coarse-to-fine zoom workflow first uses full-box VLM proposals to define a search crop and then re-localizes inside the cropped region. The metadata-constrained policy for the controlled dataset uses the same image-based scoring family applied to translation-only proposals, with the policy variant selected on the controlled pairs themselves rather than transferred from challenging.

\begin{table}[!ht]
  \centering
  \footnotesize
  \caption{Proposal-and-verify policy details by dataset and setting, corresponding to the cells in Table~\ref{tab:search_results}. Policy variant names refer to the internal experiment registry and are preserved for reproducibility.}
  \label{tab:appendix_proposal_verify_details}
  \begin{tabular}{@{}p{0.20\linewidth}p{0.28\linewidth}p{0.26\linewidth}p{0.18\linewidth}@{}}
    \toprule
    Dataset (setting) & Proposal pool & Verification and guards & Selected policy summary \\
    \midrule
    Controlled (unconstr.) & Three cached global full-box direct proposals plus three zoom proposals per pair. & Six candidates are scored after resizing crops to the XRF map. The local stage refines proposals inside cropped search regions. & Bbox-seeded zoom policy with six scored candidates per pair. \\
    Controlled (meta-constr.) & Three primary few-shot proposals per pair, scored with translation-only candidates. & Candidates are scored at the metadata-derived box size with the same scoring family used on challenging, selected on controlled. & NMI re-rank with small offset jitter. \\
    Challenging (unconstr.) & Full-box global proposals seed one zoom crop per pair, followed by three local VLM proposals. & Direct and local candidates are rescored with image evidence, and the highest-scoring candidate is retained. & Bbox-seeded zoom top-1 policy filtered to the nine small-target challenging pairs. \\
    Challenging (meta-constr.) & Three primary few-shot proposals plus three supplemental zero-shot proposals per pair. & The primary pool is scored first. A supplemental proposal can override only if it improves the score and overlaps the primary top candidate by at least 0.5 IoU. Templates with area below 60000 pixels use the NMI fallback. & Edge-guard policy with seven primary-only selections and two supplemental overrides. \\
    Leave-one-pair-out & Candidate generation is fixed; the scoring policy is selected using all pairs except the held-out pair and applied once to the held-out pair. & Held-out estimates equal the full estimates in the two primary policy-selection checks. & Challenging metadata-constrained mean 0.35; controlled metadata-constrained mean 0.20. \\
    \bottomrule
  \end{tabular}
\end{table}

On the challenging dataset under metadata-constrained search, the two supplemental overrides occurred for \texttt{2xfm\_0009} and \texttt{2xfm\_0010}, both large enough for edge-chamfer scoring. The two smallest templates, \texttt{2xfm\_0038} and \texttt{2xfm\_0042}, used the NMI fallback because edge maps were too sparse to provide a stable score. This behavior matches the intended role of the guard rules: allow candidate diversity when spatially compatible evidence supports it, but avoid unconstrained jumps to unrelated image regions. Candidate generation, the 0.5-IoU guard, and the 60{,}000-pixel area threshold are fixed configuration. Only the scoring rule is re-selected in the leave-one-pair-out analysis.

\FloatBarrier

\section{Full Numerical Results}
\label{app:full_results}

Tables~\ref{tab:appendix_summary_statistics_controlled} and~\ref{tab:appendix_summary_statistics_challenging} report mean, median, and bootstrap confidence intervals for all main methods, split by dataset, with both settings as columns. Table~\ref{tab:appendix_pair_results} gives the pair-level IoU values for the full 2-by-2 result matrix. Table~\ref{tab:appendix_statistical_tests} gives paired tests for proposal-and-verify against direct prompting and the image-center prior in every dataset--setting cell.

Key paired comparisons use exact one-sided sign-flip permutation tests. For paired differences $d_i$ and observed mean difference $\bar{d}$, the test enumerates all sign assignments $\sigma_i \in \{-1,1\}$ and computes
\[
  p = 2^{-n}\sum_{\sigma}
  \mathbb{1}\!\left[
  \frac{1}{n}\sum_i \sigma_i d_i \geq \bar{d}
  \right].
\]
This test is appropriate for small paired samples because it does not assume normality. Where sample size leaves a comparison underpowered, we report this explicitly. To assess sensitivity of the proposal-and-verify scoring-rule selection, we used leave-one-pair-out cross-validation: for each held-out pair the scoring rule was selected on the remaining pairs and applied to the held-out pair, so instability in the selected rule or a drop in held-out performance would indicate sensitivity to individual pairs.

\begin{table}[!ht]
  \centering
  \scriptsize
  \caption{Summary statistics for localization IoU on the controlled dataset ($n{=}5$), both settings. Confidence intervals are bootstrap 95\% intervals over image pairs. The unconstrained classical entry uses multi-scale gradient+NCC search, and the metadata-constrained entry uses a fixed template at the ground-truth box size. The unconstrained direct VLM entry asks for a full four-parameter box, and the metadata-constrained entry asks for a normalized top-left position over the valid placement range. The unconstrained proposal-and-verify entry uses a coarse-to-fine zoom workflow, and the metadata-constrained entry uses the best image-based scoring policy fitted on the controlled dataset itself. As a transfer diagnostic, the metadata-constrained policy selected on challenging yields 0.11 [0.03, 0.20] when applied to controlled.}
  \label{tab:appendix_summary_statistics_controlled}
  \resizebox{\linewidth}{!}{%
  \begin{tabular}{@{}lcc@{}}
    \toprule
    Method & Unconstrained mean (median; 95\% CI) & Metadata-constrained mean (median; 95\% CI) \\
    \midrule
    Image-center prior & 0.00 (0.00; 0.00--0.00) & 0.00 (0.00; 0.00--0.00) \\
    Random-box prior & 0.01 (0.01; 0.01--0.02) & 0.01 (0.01; 0.01--0.02) \\
    Constant prior & 0.11 (0.00; 0.00--0.33) & 0.11 (0.00; 0.00--0.33) \\
    DINOv2 dense features & 0.06 (0.00; 0.00--0.14) & 0.06 (0.00; 0.00--0.14) \\
    multiGradICON pretrained & 0.00 (0.00; 0.00--0.00) & 0.00 (0.00; 0.00--0.00) \\
    Classical template matching & 0.12 (0.00; 0.00--0.36) & 0.92 (0.90; 0.89--0.95) \\
    Direct VLM, 3 examples & 0.05 (0.01; 0.00--0.09) & 0.10 (0.08; 0.04--0.16) \\
    Proposal-and-verify & 0.09 (0.00; 0.00--0.19) & 0.20 (0.21; 0.05--0.35) \\
    \bottomrule
  \end{tabular}%
  }
\end{table}

\begin{table}[!ht]
  \centering
  \scriptsize
  \caption{Summary statistics for localization IoU on the challenging dataset ($n{=}9$ small-target pairs), both settings. Confidence intervals are bootstrap 95\% intervals over image pairs. The unconstrained classical entry uses multi-scale gradient+NCC search, and the metadata-constrained entry uses a fixed template at the probe-metadata-derived box size. The unconstrained direct VLM entry asks for a full four-parameter box, and the metadata-constrained entry asks for a normalized top-left position over the valid placement range. The unconstrained proposal-and-verify entry uses a coarse-to-fine zoom workflow, and the metadata-constrained entry uses the edge-chamfer scoring policy with NMI fallback selected via leave-one-pair-out on the challenging dataset.}
  \label{tab:appendix_summary_statistics_challenging}
  \resizebox{\linewidth}{!}{%
  \begin{tabular}{@{}lcc@{}}
    \toprule
    Method & Unconstrained mean (median; 95\% CI) & Metadata-constrained mean (median; 95\% CI) \\
    \midrule
    Image-center prior & 0.25 (0.07; 0.06--0.48) & 0.25 (0.07; 0.06--0.48) \\
    Random-box prior & 0.06 (0.02; 0.01--0.13) & 0.06 (0.02; 0.01--0.13) \\
    Constant prior & 0.20 (0.09; 0.03--0.42) & 0.20 (0.09; 0.03--0.42) \\
    DINOv2 dense features & 0.20 (0.00; 0.00--0.40) & 0.20 (0.00; 0.00--0.40) \\
    multiGradICON pretrained & 0.19 (0.00; 0.02--0.37) & 0.19 (0.00; 0.02--0.37) \\
    Classical template matching & 0.00 (0.00; 0.00--0.00) & 0.01 (0.00; 0.00--0.02) \\
    Direct VLM, 3 examples & 0.11 (0.09; 0.04--0.18) & 0.19 (0.14; 0.06--0.32) \\
    Proposal-and-verify & 0.19 (0.10; 0.05--0.39) & 0.35 (0.37; 0.14--0.58) \\
    \bottomrule
  \end{tabular}%
  }
\end{table}

\begin{table}[!ht]
  \centering
  \scriptsize
  \caption{Pair-level IoU values for evaluated methods, with one row per pair from each dataset. Geometric controls and pretrained baselines are setting-invariant under the evaluated box-size convention. Classical template matching, direct VLM, and proposal-and-verify are reported separately for unconstrained and metadata-constrained search. Setting and policy choices per cell are documented in Tables~\ref{tab:appendix_summary_statistics_controlled}--\ref{tab:appendix_summary_statistics_challenging} and Table~\ref{tab:appendix_proposal_verify_details}.}
  \label{tab:appendix_pair_results}
  \resizebox{\linewidth}{!}{%
\begin{tabular}{@{}llrrrrrrrrrrrr@{}}
\toprule
Dataset & Pair & Area & Center & Random & Const. & DINOv2 & mGICON & \multicolumn{3}{c}{Unconstrained} & \multicolumn{3}{c}{Metadata-constrained} \\
\cmidrule(lr){9-11}\cmidrule(l){12-14}
 & & (\%) & & & & & & Class. & Direct & P\&V & Class. & Direct & P\&V \\
\midrule
Controlled & \texttt{60fb54} & 1.8 & 0.00 & 0.02 & 0.56 & 0.21 & 0.00 & 0.60 & 0.00 & 0.00 & 0.97 & 0.00 & 0.00 \\
 & \texttt{6dfe75} & 1.1 & 0.00 & 0.01 & 0.00 & 0.00 & 0.00 & 0.00 & 0.11 & 0.00 & 0.89 & 0.13 & 0.47 \\
 & \texttt{85e863} & 1.6 & 0.00 & 0.01 & 0.00 & 0.10 & 0.00 & 0.00 & 0.00 & 0.19 & 0.90 & 0.07 & 0.21 \\
 & \texttt{97acdf} & 1.2 & 0.00 & 0.01 & 0.00 & 0.00 & 0.00 & 0.00 & 0.01 & 0.00 & 0.97 & 0.08 & 0.02 \\
 & \texttt{e1c7e5} & 1.5 & 0.00 & 0.01 & 0.00 & 0.00 & 0.00 & 0.00 & 0.11 & 0.25 & 0.89 & 0.21 & 0.30 \\
\addlinespace[0.25em]
Challenging & \texttt{2xfm\_0009} & 19.3 & 0.74 & 0.31 & 0.89 & 0.59 & 0.74 & 0.00 & 0.08 & 0.10 & 0.05 & 0.49 & 0.85 \\
 & \texttt{2xfm\_0010} & 9.1 & 0.41 & 0.12 & 0.57 & 0.38 & 0.38 & 0.00 & 0.00 & 0.00 & 0.00 & 0.47 & 0.87 \\
 & \texttt{2xfm\_0012} & 0.5 & 0.00 & 0.00 & 0.00 & 0.00 & 0.00 & 0.00 & 0.00 & 0.00 & 0.00 & 0.00 & 0.00 \\
 & \texttt{2xfm\_0013} & 0.3 & 0.87 & 0.00 & 0.13 & 0.00 & 0.50 & 0.00 & 0.00 & 0.00 & 0.00 & 0.00 & 0.00 \\
 & \texttt{2xfm\_0016} & 2.5 & 0.00 & 0.02 & 0.00 & 0.00 & 0.00 & 0.00 & 0.09 & 0.05 & 0.00 & 0.00 & 0.00 \\
 & \texttt{2xfm\_0017} & 2.2 & 0.20 & 0.02 & 0.09 & 0.00 & 0.00 & 0.00 & 0.10 & 0.10 & 0.00 & 0.16 & 0.37 \\
 & \texttt{2xfm\_0022} & 4.8 & 0.07 & 0.05 & 0.12 & 0.81 & 0.07 & 0.00 & 0.18 & 0.20 & 0.00 & 0.42 & 0.63 \\
 & \texttt{2xfm\_0038} & 0.2 & 0.00 & 0.00 & 0.00 & 0.00 & 0.00 & 0.00 & 0.19 & 0.27 & 0.00 & 0.00 & 0.00 \\
 & \texttt{2xfm\_0042} & 0.2 & 0.00 & 0.00 & 0.00 & 0.00 & 0.00 & 0.00 & 0.32 & 0.95 & 0.00 & 0.14 & 0.43 \\
\bottomrule
\end{tabular}%
}

\end{table}

\begin{figure}[!ht]
  \centering
  \includegraphics[width=\linewidth]{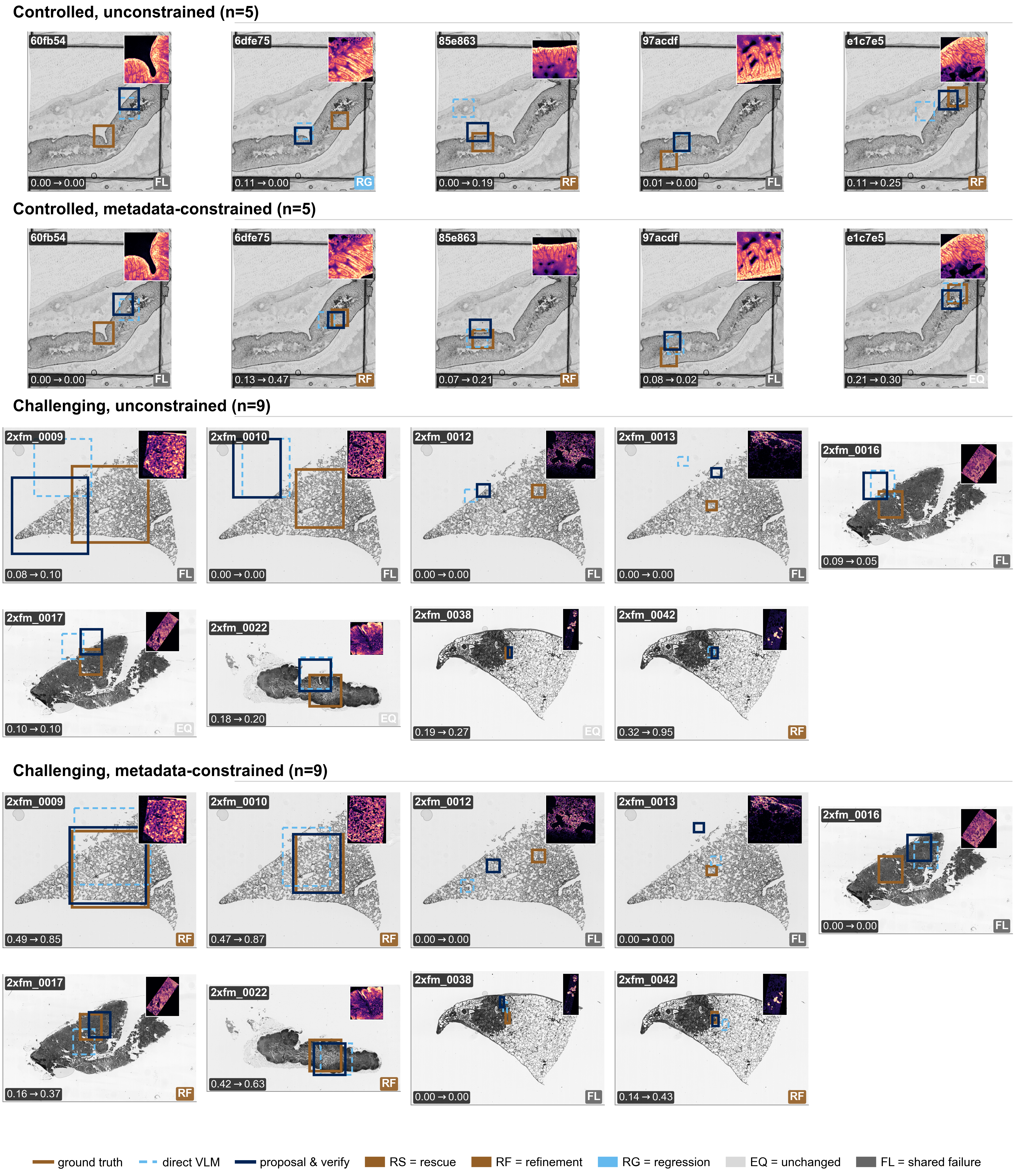}
  \caption{Qualitative atlas for the evaluated pairs and both search settings, visualizing the direct VLM and proposal-and-verify columns of Table~\ref{tab:appendix_pair_results}. Each optical reference shows ground truth, direct VLM prediction, and proposal-and-verify prediction using the same color convention as Fig.~\ref{fig:qualitative_examples}. Outcome codes distinguish rescue (RS), refinement (RF), regression (RG), unchanged (EQ), and shared failure (FL) cases, matching the legend rendered on the figure.}
  \label{fig:appendix_qualitative_atlas}
\end{figure}

\begin{table}[!ht]
  \centering
  \footnotesize
  \caption{Statistical comparisons for the proposal-and-verify workflow in all four dataset--setting cells. Paired tests are exact one-sided sign-flip tests over image pairs.}
  \label{tab:appendix_statistical_tests}
  \begin{tabular}{@{}llrrrr@{}}
\toprule
Dataset & Setting & Direct & P\&V & $p$(P\&V $>$ direct) & $p$(P\&V $>$ center) \\
\midrule
Controlled & unconstrained & 0.05 & 0.09 & $0.250$ & $0.250$ \\
Controlled & metadata-constrained & 0.10 & 0.20 & $0.125$ & $0.062$ \\
Challenging & unconstrained & 0.11 & 0.19 & $0.125$ & $0.648$ \\
Challenging & metadata-constrained & 0.19 & 0.35 & $0.031$ & $0.266$ \\
\bottomrule
\end{tabular}

\end{table}

\FloatBarrier

\section*{Acknowledgments}
This research used resources of the Advanced Photon Source, a U.S.~Department of Energy (DOE) Office of Science user facility at Argonne National Laboratory. The authors thank the staff of APS beamlines 2-ID-D and 2-ID-E, particularly Evan Maxey and Olga Antipova, for support with the XRF measurements.

\section*{Funding}
This research is based on work supported by the U.S. DOE Office of Science-Basic Energy Sciences, under Contract No.~DE-AC02-06CH11357. The authors acknowledge funding from DOE Office of Science-Basic Energy Sciences, award 0000283133. In addition, a portion of the research support for this project came from the National Institutes of Health NIH/NIAID award 1P01AI165380-01.

\section*{Data and Code Availability}
The data and code supporting the findings of this study will be released upon publication.

\bibliographystyle{unsrt}
\bibliography{references}

\end{document}